\documentclass[sn-mathphys-ay, iicol]{sn-jnl}

\usepackage{graphicx}%
\usepackage{multirow}%
\usepackage{amsmath,amssymb,amsfonts}%
\DeclareMathOperator{\sign}{Sign}
\DeclareMathOperator{\argmax}{argmax}
\usepackage[title]{appendix}%
\usepackage{xcolor}%
\usepackage{textcomp}%
\usepackage{manyfoot}%
\usepackage{booktabs}%
\usepackage{enumitem}%

\begin{document}

\title[Article Title]{Exploring Biologically Inspired Mechanisms of Adversarial Robustness}


\author[1]{\fnm{Konstantin} \sur{Holzhausen}}\email{holzhausen.research@gmail.com}

\author[1]{\fnm{Mia} \sur{Merlid}}

\author[1]{\fnm{H\aa kon Olav} \sur{Torvik}}

\author[1]{\fnm{Anders} \sur{Malthe-S\o renssen}}

\author*[1, 2]{\fnm{Mikkel Elle} \sur{Lepper\o d}}\email{mikkel@simula.no}

\affil[1]{\orgdiv{Department of Physics}, \orgname{University of Oslo}, \orgaddress{\street{Sem S\ae lands vei 24, Fysikkbygningen}, \city{Oslo}, \postcode{0371}, \country{Norway}}}

\affil*[2]{\orgdiv{Department of Numerical Analysis and Scientific Computing}, \orgname{Simula Reserach Laboratory}, \orgaddress{\street{Kristian Augusts gate 23}, \city{Oslo}, \postcode{0164}, \country{Norway}}}



\abstract{
Backpropagation-optimized artificial neural networks, while precise, lack robustness, leading to unforeseen behaviors that affect their safety. 
Biological neural systems do solve some of these issues already.
Unlike artificial models, biological neurons adjust connectivity based on neighboring cell activity.
Understanding the biological mechanisms of robustness can pave the way towards building trustworthy and safe systems. 
Robustness in neural representations is hypothesized to correlate with the smoothness of the encoding manifold. 
Recent work suggests power law covariance spectra, which were observed studying the primary visual cortex of mice, to be indicative of a balanced trade-off between accuracy and robustness in representations.
Here, we show that unsupervised local learning models with winner takes all dynamics learn such power law representations, providing upcoming studies a mechanistic model with that characteristic.
Our research aims to understand the interplay between geometry, spectral properties, robustness, and expressivity in neural representations.
Hence, we study the link between representation smoothness and spectrum by using weight, Jacobian and spectral regularization while assessing performance and adversarial robustness. 
Our work serves as a foundation for future research into the mechanisms underlying power law spectra and optimally smooth encodings in both biological and artificial systems. 
The insights gained may elucidate the mechanisms that realize robust neural networks in mammalian brains and inform the development of more stable and reliable artificial systems.
}




\maketitle

\section{Introduction}\label{intro}

Research on Artificial Intelligence (AI) has made tremendous progress within recent decades. 
To a large extent, this success is due to biologically inspired artificial neural networks (ANN) with vast parameter spaces. 
Convolutional Neural Networks (CNNs) constitute a prominent example.
Mimicking the morphology of the visual cortex, they have revolutionized the field of image analysis \citep{Krizhevsky.2012}.
Today, the majority of ANNs are trained supervised using backpropagation.

Despite achieving high accuracy, backpropagation optimized ANNs are unstable with regard to changes in their input \citep{Antun.2020}.
For example, unpredictable changes in output are caused by random noise or adversarial examples on the input \citep{Goodfellow.2015}, \citep{Dezfooli.2017}.
Instabilities can lead to unexpected model behaviors with direct consequences for the applicability of AI technology.
In cancer recognition, for example, hardly visible changes in images of moles can cause diagnostic tools to change their rating from benign to malignant \citep{Finlayson.2019} at the expense of the patient's health.
Not only does this example illustrate direct individual implications, but it also questions the reliability of such systems and can be a problem of societal impact.

Compared to humans in a feed forward setting, machine learners are significantly less robustness against black box attacks \citep{Geirhos.2018}.
This type of divergence is hypothesized to be due to invariances in model "metamers" compared between biological and artificial neural networks \citep{Feather.2023} and perceptual straightness of visual representations \citep{Harrington.2022}.
However, black box attacks show that inferior temporal gyrus neurons in primates are more susceptible than adversarially trained networks \citep{Guo.2022}. 
Out-of-distribution (ood) generalization compared between humans and machine learners shows that the primary factors for increasing robustness are data size and architectural design \citep{Geirhos.2021}.
This example illustrates that it is promising to search for properties and mechanisms in biological systems that ANNs might benefit from and vice versa.
Biological Neural Networks (BNN) are models that emulate the nature of neural tissue beyond a connected set of neurons.
Moreover, BNNs adjust their connectivity in response to the activity patterns of neighboring neurons within the network.
In that sense, BNNs learn locally.
A prominent example of a local learning algorithm is \citeauthor{Oja.1982}'s rule \citep{Oja.1982} which is a mathematical formalization of \citeauthor{Hebb.1949}'s learning theory \citep{Hebb.1949}.
Since BNNs learn differently from ANNs, implementing principles such as local learning constitutes one potential approach to resolve robustness issues.

Recent work from \cite{Krotov.2019} introduced the idea to learn latent representations using a biologically plausible local learning rule in an otherwise backpropagation optimized model.
\cite{Grinberg.2019} demonstrated that models that feature such biologically plausible layers can at least keep up with end-to-end backpropagation trained networks in terms of accuracy.
Additionally, \cite{Patel.2020} showed similar models to be more resilient to black box attacks, such as square occlusion, than their end-to-end counterparts.
Black box attacks are perturbation methods that treat models as black boxes.
In contrast to them, white box attacks have access to the inner workings of a model and can, therefore, fool them more specifically.
These studies find that local learning yields smoother feature maps compared with their end-to-end counterparts, and conclude that to be the reason for the observed increase in robustness.

Representations on smoother manifolds are less susceptible to input perturbations, making them more robust. 
In neural networks, inputs initiate neural response patterns -- called representations -- that capture how inputs are processed.
The geometric structure of the manifold formed by all possible representations reveals insights into the network’s behavior and properties.
A smooth manifold ensures that representations transition gradually between similar inputs, reinforcing robustness by maintaining consistency.
However, smoothness can also compress distances between similar representations, making them harder to distinguish.
To improve expressivity, the model pushes representations apart while maintaining their local relation.
That separation enables better discrimination but roughens the surface.
Balancing this interplay between smoothness for robustness and representation separation for accuracy is a core challenge in representation learning.

\cite{Stringer.2019} proved that a representation's fractal dimension, which can be considered a measure of smoothness, is related to the exponent of asymptotic decay in the manifold's covariance (PCA) spectrum.
As a consequence, an optimal balance between accuracy and robustness is characterized by a close to $n^{-\alpha}$ power law decay in ordered spectral components, where $\alpha$ depends on the input's intrinsic dimension.
Interestingly, it is this power law functional relation that they also observe in the neural activities from the primary visual cortex area V1 in mice.
Not only does this result validate their argument, but it also suggests that these V1 representations are in that sense optimal.
In consequence, their study indicates that instabilities of artificial neural network models may be related to the smoothness of their representations, as compared to biological neural networks. \citep{Stringer.2019}
Following this, \cite{Nassar.2020} introduced a power law spectral regularization term to enforce their image classifiers to favor power law representations in their hidden layers under supervised learning.
In agreement with Stringer et al., they found representations following a power law to be more robust in Multi-Layer Perceptron Models (MLPs) and CNNs.
However, their results are solely empirical and the underlying mechanisms are not understood.

Instead of relying on empirical evidence linking the spectrum to robustness, one can optimize for smooth representations directly.
Assuming the representation manifold to be locally differentiable, the norm of its Jacobian constitutes a valid local measure of change and, hence, smoothness.
Thus, bounding the Jacobian's magnitude provides a regularization mechanism to achieve smoother manifolds \citep{Varga.2018}.
\cite{Hoffman.2019} found the decision landscapes of Jacobian regularized classifiers to change less abruptly, with smoother boundaries increasing resilience to adversarial attacks. 
Their algorithm constrains hidden representations in favor of differential smoothness and robustness efficiently.

In summary, it appears that a representation's geometry, its spectrum and robustness are closely interrelated.
Although some links are well understood, in general their mutual dependence remains unclear.
Since Krotov and Hopfield's learning rule is directly related to the physiological learning processes, to test whether it reproduces \citeauthor{Stringer.2019}'s spectral property seems obvious.
Being a mechanistic model makes it in principle an ideal candidate device for understanding these connections in the future since its connectivity dynamics are explicitly stated.
In this paper, we study the cross relations between the power spectral decay, geometric properties, robustness and expressivity (model performance).
At first, we test \citeauthor{Krotov.2019}'s model for adversarial robustness with respect to random corruption and white box attacks.
Based on the results in \cite{Patel.2020}, we expect it to be more robust compared to end-to-end backpropagation trained shallow models.
In an attempt to understand the underlying mechanism, we probe the representation's properties.
At first, we examine whether the model reproduces spectral decays conforming to a power law.
Using the mentioned regularization methods allows to specifically constrain the optimization for smooth or power law compliant representations.
With this, we study the mutual implications of the latent characteristics in a systematic manner.

We will provide a short overview of our architectural choices, Krotov and Hopfield's local learning model, optimization and regularization methods as well as dataset and hyperparameter choices in the following methods section.
Thereafter, we present our results in a logical order followed by a discussion.

\section{Methods}\label{meth}
\subsection*{Architectural choices}
To study the implications of structural properties in the hidden representation, we assess relative model performance and constrain the structure of our neural network model to an Encoder-Decoder architecture.
This choice justifies simpler function classes to control parameters and restricts effects due to large model complexity.
Because the regularizers are architecture-agnostic beyond the existence of a hidden layer, compatibility concerning \citeauthor{Krotov.2019}'s hybrid model is the limiting factor.
Consequently, we choose a Multi-Layer Perceptron model, similar to that in \cite{Krotov.2019}
\begin{equation}
    \begin{aligned}
        h(x) &= W \, x \\
        \hat{h}(h) &= \text{ReLU}(h)^{n} \\
        y(\hat{h}) &= A\, \hat{h} + b \quad \text{.}
    \end{aligned}
    \label{eq:forwardpass}
\end{equation}
Here, $x$ denotes a flattened single image.

\subsection*{Representation spectrum}
\citeauthor{Stringer.2019}'s theory makes statements about the functional relationship between eigenvalues and their index in the ordered spectrum of principal components in the set of the model's representations.
Because principal components correspond to the eigenvectors of the respective covariance matrix $\text{Cov}(h, h)$, we study the covariance spectrum $\{\lambda_{n}\}_{1 \leq n \leq N}$ in descending order:~$\lambda_{1} \geq \lambda_{2} \geq \dots \geq \lambda_{N}$.
According to \cite{Stringer.2019}, optimal encodings follow a power law
\begin{equation}
    \lambda_{n} = \lambda_{1} \, n^{-\alpha}
\end{equation}
with an exponent $\alpha > 1$.
To detect power laws, we make use of their scale invariance property.
For any fixed exponent, scaling transformations of the kind $n \mapsto a\, n$ induce
\begin{equation}
    \frac{\lambda_{n}}{\lambda_{1}} \mapsto \frac{\lambda_{a\, n}}{\lambda_{a}} = \frac{\lambda_{1}}{\lambda_{a}} \, \left( a\, n \right)^{-\alpha} = n^{-\alpha} \text{,}
    \label{eq:PowerLawScaling}
\end{equation}
which follows the same power law in $n$.
Thus, the functional relationship in $\tilde{\lambda}_{n} = \lambda_{n} / \lambda_{1}$ is scale invariant, and is therefore comparable between different models at arbitrary scales.
As a result, we measure the decay exponent in the normalized spectrum $\tilde{\lambda}_{n}$.
On the flip side, this also means that power laws remain maintained, even under changing the dimensionality of representations, offering a qualitative test to discriminate real power laws from other similar relations.

We restrict our quantitative analysis of the power law to identifying the exponent $\alpha$ using linear regression on its double logarithmic representation of normalized eigenvalues.
At this point we stress the important difference between our notion of a power law and what is widely perceived.
Primarily, power laws refer to probability distributions or their density functions. 
In both cases, it is best to estimate parameters leveraging the vast amounts of statistical methods that exist.
Here, however, by power law we refer to a functional relationship, whose parameters can be inferred by regression methods.
In a double logarithmic plot, power laws appear as linear relations
\begin{equation}
    \log{\tilde{\lambda}_{n}} = -\alpha\, \log{n} \quad \text{,}
\end{equation}
with former exponents corresponding to slopes.
Therefore, we use linear regression to identify $\alpha$ and its error in the double logarithmic representation of the ordered and normalized spectrum.
Because statistical tests for linear regression focus on monotonicity hypotheses, which are trivially met in ordered spectra, we renounce analyses regarding our estimates' significance.
Since we expect to see boundary effects, we will do this analysis in regions away from the boundary.

\subsection*{Unsupervised training and Post-processing}\label{unsuplearn}
\cite{Krotov.2019} introduce a biologically inspired dynamic learning rule to learn latent representations of images in an unsupervised scheme.
In a network of hidden units $\{h_{i}\}$, the synaptic strengths $S_{i, j}$ to the input units $\{x_{j}\}$ are updated based on approximations of renowned neuroplastic mechanisms.
A batch parallel version reads
\begin{equation}
    \Delta S_{i, j} = \eta_{L} \, \mathbb{E}_{x \in \mathcal{B}} \left[ g \left( h_{i}(x) \right)\, \left( x_{j} - h_{i}(x)\, S_{i, j} \right) \right]
    \label{eq:llrule}
\end{equation}
where the function 
\begin{equation}
    g(h_{i}) = \begin{cases}
        1 &\Leftrightarrow \quad h_{i} = \max{\left[ h \right]} \\
        - \delta  &\Leftrightarrow \quad h_{i} = \max^{k}{\left[h\right]} \\
        0 & \text{else}
    \end{cases}
    \label{eq:rankg}
\end{equation}
realizes a Winner-Take-All as well as an inhibition mechanism.
The synaptic strengths $S_{i, j}$ relate to the weights $W$ in Equation~\eqref{eq:forwardpass} by $W_{i, j} = S_{i, j}\, | S_{i, j} |^{p-2}$, where the parameter $p$ sets the measure of distances between image vectors.
In Equation~\eqref{eq:rankg}, $\max^{k}\left[ \cdot \right]$ denotes the $k$-th maximum in the set of entries of the vector $h$ and $\delta$ the effective inhibition strength.
We choose the values of those hyperparameters including the learning rate $\eta_{L}$ consistently with the original model.
Under these conditions, the model learns prototypic representations and some other features of the data set.
Whilst this outlines the mechanistic rule according to which latent representations form, the final classification layer is trained using supervised learning with backpropagation.

\begin{figure}[ht]
\begin{center}
\includegraphics[width=\linewidth]{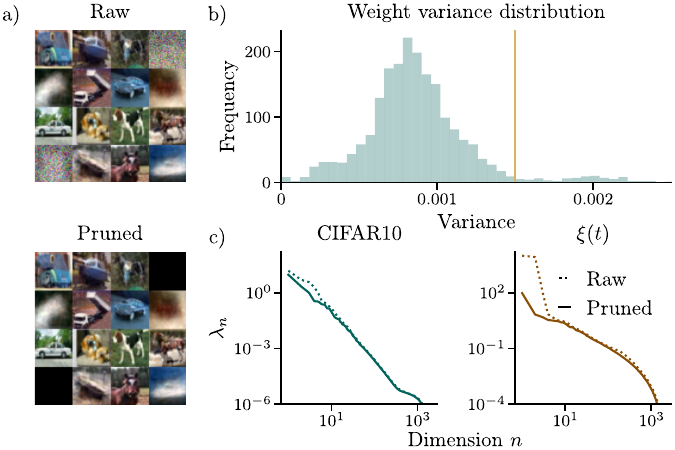}
\end{center}
\caption{
Synapse and representation characteristics after unsupervised training with the local learning rule. 
\textbf{a)}:~Synaptic connections presented as images.
Some entry blocks appear to resemble noise (Raw).
We prune noisy contributions to improve spectral properties (Pruned).
\textbf{b)}:~Distribution of total variances per image block in $S$. 
The distribution is bimodal with the major mode centered below $0.001$ and the minor mode around $0.002$. 
Pruning the higher variance contributions by setting a cutoff threshold at $0.015$ ablates noisy images in $S$ (see panel \textbf{a)}), and eliminates the initial drop in the representation's covariance spectrum (see panel \textbf{b)}).
The latter results in comparably clean power relations between the eigenvalues of the first components.
\textbf{c)}:~Ordered covariance spectra of the representations corresponding to both, the original (Raw) and ablated (Pruned) set of synapses, subjected to CIFAR10 (left panel) or Gaussian noise (right panel).
}
\label{fig:llsynapses}
\end{figure}

To avoid unwanted side effects, we ablated synapses for which we had reason to believe they have not fully converged.
Figure~\ref{fig:llsynapses}~a) shows selected synaptic weights in $S$ after unsupervised training with CIFAR10 images.
Each image block represents the weights $S_{i, \cdot}$ projecting inputs onto hidden units.
While the model captures prototypic features, some blocks appear random, suggesting they lack meaningful information due to limited training. 
These noisy synapses contribute high variance, as seen in the distribution of per-image variances in Figure~\ref{fig:llsynapses}~b).
This distribution exhibits two peaks: a major mode near $0.001$ and a secondary one around $0.002$.
Assuming the latter represents unconverged synapses, we removed connections with variances exceeding $0.0015$, eliminating the noisy blocks (Figure\ref{fig:llsynapses}~a)~Pruned).

Figure~\ref{fig:llsynapses}~c) compares the ordered covariance spectra of latent representations before and after ablation for CIFAR10 test images (left) and random noise inputs (right). 
Both profiles suggest a power-law, but this visual assessment is insufficient for definitive confirmation.
A comprehensive analysis is presented in Section~\ref{res}.
In the unpruned model, the noise-input spectrum shows a sharp drop after the first two eigenvalues, which is mirrored in the CIFAR10 spectrum. 
Ablation removes these artifacts, yielding more coherent spectra and supporting the hypothesis the additional variance contributions were caused by the ablated noisy synapses.
Based on these findings, the pruned model was selected for further analysis.

Automation of the described post-processing procedure is implemented as a two stage process.
First, we assess whether the unsupervised learning dynamics have reached a steady state, indicating that the model has learned something.
A model is considered sufficiently converged if more than $10 \%$ of feature weight vectors stabilize near $\left| S_{i, \cdot} \right|^{p} \simeq R = 1.0$ within a tolerance of $10^{-2}$. 
Additionally, our empirical observations suggest the mean feature weight to exceed $\mathbb{E} \left[ S_{i, \cdot} \right] > -R / 2$, avoiding convergence to an undesired steady state such as $S_{i, j} \equiv -1$.
Models failing these criteria are rejected.
Converged models exhibit multi-modal variance distributions, as seen in Figure~\ref{fig:llsynapses}~b).
Given that variances are strictly positive, we model their distribution using a mixture of up to four Log-normal components, creating a hierarchy of parametric models with degrees of freedom differing by $df = 3$.
Using a likelihood ratio test, we perform model selection based on rejection of the less complex model ($p < 0.01$).
If the chosen distribution is uni-modal, no pruning is performed.
Otherwise, if multiple components are detected, high-variance contributions, attributed empirically to noisy, non-converged synapses, are pruned with the threshold set at half the distance between the means of the two highest-variance components.
In summary, this process automates model selection based on dynamic learning parameters and variance distribution modeling, ensuring robust outcomes.

\subsection*{Supervised learning, loss and regularizations}
All supervised training, which includes learning only the decoder in Krotov and Hopfield's hybrid model (KH) and all weights in the Single Hidden Layer Perceptron (SHLP), was based on optimizing the Cross Entropy Loss using the Adam optimizer and mini batches consisting of $1000$ examples each.
Besides the pure SHLP, we complemented the total loss by adding regularization terms to achieve desired properties in the hidden layer.
To study the implications of smoother encodings, we used L2 and \citeauthor{Hoffman.2019}'s Jacobian regularization with $n_{\text{proj}}=3$ projections, but only on hidden features.
Moreover, we spectrally regularizered the hidden representations for a $n^{-1}$ spectrum using \citeauthor{Nassar.2020}'s method.
For the results in this paper, we used the CIFAR10 datasets for training and testing.

In terms of structural and spectral properties of the hidden, or rather latent, representations, we expect both, L2 and Jacobian regularization, to have a similar effect. 
In the upcoming subsection, we lay down our argument more comprehensively.
However, since we measure robustness with regard to the model's prediction, which involves decoding of the hidden representation, they might affect our robustness measures differently.
We suspect such also because local and global bounding might affect decoding differently.

\subsection*{Similarity of L2 and Jacobian regularization}\label{regularizations}
In terms of spectral implications, we should expect weight based regularization methods to have a similar effect, whether it is Jacobian or L2 regularization.
The notions of robustness and continuity are closely related in the context of static models or functions.
Therefore, Stringer's statement, about smoother representations accounting for more robust models, intuitively makes sense.
With robustness, we are generally interested in how comparably small perturbations in the input $x \mapsto x + p$ locally translate into changes in the output.
With \cite{Stringer.2019} in mind, we are particularly interested in changes in the latent or hidden representation $\hat{h}(x) \mapsto \hat{h}(x + p)$.
Assuming local differentiability allows us to locally quantify how perturbations translate by applying Taylor expansion to the perturbed feature.
\begin{equation}
    \begin{aligned}
        \left\| \hat{h}(x + p) - \hat{h}(x) \right\| =& \left\| \hat{h}(x) + D \hat{h}(x)\cdot p + \right. \\
        & \left. + \dots - \hat{h}(x) \right\| \\
        \leq \left\| D \hat{h}(x) \right\|& \, \|p\| + \mathcal{O}(\|p\|^{2})
    \end{aligned}
    \label{eq:taylorexp}
\end{equation}
Comparably low perturbations in the input scale with a factor of $\|D \hat{h}(x)\|$ in the representations.
Thus, bounding the Jacobian $J_{ij} = \partial_{x_j}\, \hat{h}_{i}(x) = \left( D \hat{h}(x) \right)_{i,j}$ yields local stability with respect to small perturbations.
Because structurally $\hat{h}(x) = \sigma(W\, x)$, the norm of the weights directly relates to the norm of the Jacobian.
Jacobian regularization, according to \cite{Hoffman.2019}, minimizes the Frobenius norm of the Jacobian tensor $\left\| J \right\|_{\mathcal{F}} = \sum_{i,j} |J_{i, j}|^{2}$.
Using Equation~\ref{eq:forwardpass}, one gets
\begin{equation}
    \begin{aligned}
        \left| J_{i, j} \right|^{2} =& \left| \sigma^{\prime}\left[ (Wx)_{i} \right]\, \partial_{x_{j}}\, \sum_{k} W_{i,k}\, x_{k} \right|^{2}\\ 
        =& \left| \sigma^{\prime}\left[ (Wx)_{i} \right] \right|^{2}\, \left| W_{i,j}\right|^{2} \quad \text{.}
    \end{aligned}
\end{equation}
In consequence, we can express the Frobenius norm as
\begin{equation}
    \begin{aligned}
        \sum_{i}& \left( \sum_{j} \left| J_{i, j} \right|^{2} \right)= \\
        &=\sum_{j}\, \sum_{i} \left| \sigma^{\prime}\left[ (Wx)_{i} \right] \right|^{2} \left| W_{i,j} \right|^{2}\\
        &= \sum_{j}\, \left| \sigma^{\prime}\left[ Wx \right] \right|^{2} \cdot \left| W_{\cdot,j} \right|^{2} \quad \text{.}
    \end{aligned}
\end{equation}
Using the Cauchy-Schwartz inequality allows one to rewrite the dot product.
\begin{equation}
    \begin{aligned}
        \sum_{j}\,& \left| \sigma^{\prime}\left[ Wx \right] \right|^{2} \cdot \left| W_{\cdot,j} \right|^{2} \leq \\
        &\leq \sum_{j}\, \sqrt{\sum_{i} \left| \sigma^{\prime}\left( Wx \right)_{i} \right|^{4}}\, \sqrt{\sum_{i} \left| W_{i, j} \right|^{4}}
    \end{aligned}
    \label{eq:cauchyschwarzt}
\end{equation}
Because
\begin{equation}
    \sum_{i} \left| \sigma^{\prime}\left( Wx \right)_{i} \right|^{4} \leq \left( \sum_{i} \left| \sigma^{\prime}\left( Wx \right)_{i} \right|^{2} \right)^{2}
\end{equation}
which also holds for
\begin{equation}
    \sum_{i} \left| W_{i,j} \right|^{4} \leq \left( \sum_{i} \left| W_{i,j} \right|^{2} \right)^{2} \quad \text{,}
\end{equation}
Equation~\ref{eq:cauchyschwarzt} reads
\begin{equation}
    \begin{aligned}
        \left\| J \right\|_{\mathcal{F}} \leq& \sum_{i} \left| \sigma^{\prime}\left[ \left( Wx \right)_{i} \right] \right|^{2} \, \sum_{i, j} \left| W_{i,j} \right|^{2} \\
        =& \left\| \sigma^{\prime}\left[ Wx \right] \right\|^{2}_{2} \, \sum_{j}\, \left\| W_{\cdot, j} \right\|^{2}_{2}\\
        =& \left\| \sigma^{\prime}\left[ Wx \right] \right\|^{2}_{2}\, \left\| W \right\|_{\mathcal{F}} \quad \text{.}
    \end{aligned}
\end{equation}
Thus, if $\sigma^{\prime}\left[ Wx \right]$ is locally bounded, L2 weight regularization implies Jacobian regularization in that region.
Consequently, L2 and Jacobian regularization have locally the same effect on robustness.
Since in our case $\sigma = \text{ReLU}^{n}(\cdot)$, the activation function is not differentiable at singular sets.

\subsection*{Perturbation experiments} 
We tested model robustness against random perturbations and adversarial attacks.
For random perturbations, test images $x$ were perturbed in random directions $\xi \sim \mathcal{N}(0, \mathbb{I})$ 
\begin{equation}
    x_{\xi} = x + \epsilon\; \xi / \|\xi\|
\end{equation}
by a length $\epsilon > 0$.
Next to random perturbations, we tested robustness against adversarial attacks, including the Fast Gradient Sign Method (FGSM) and Projected Gradient Descent (PGD).
FGSM moves $x$ in the direction of maximum loss increase
\begin{equation}
    x_{\text{FGSM}} = x + \epsilon\; \sign\left[ \nabla_{x} \mathcal{L}\left( y(x) \right) \right]
\end{equation} by magnitude $\epsilon$ once \citep{Goodfellow.2015}.
Here, $\mathcal{L}\left(y(x) \right)$ denotes the loss function from the supervised training which quantifies the discrepancy between model prediction $y(x)$ and label.
PGD \citep{Kurakin.2017, Madry.2018} iterates FGSM $N = 10$ times 
\begin{equation}
    \begin{aligned}
        x_{s} &= \mathcal{P}_{\mathcal{B}_{\epsilon}(x)} \left[ x_{s - 1} \right. +\\
        &\quad + \left. \delta \sign \left[ \nabla_{x_{s-1}} \mathcal{L}\left( y(x_{s-1}) \right) \right] \right]
    \end{aligned}
\end{equation}
and projects the results into the $\epsilon$ ball around the original image $x_{0} = x$.
As a result, $x_{\text{PGD}} = x_{N}$ is an approximation of the most confusing input example that is at most by $\epsilon$ apart from $x$.
For PGD, we chose step lengths $\delta = 10^{-1} \epsilon$.

To measure model performance under these perturbations, we captured two metrics: relative accuracy and critical distance.
Because each model achieves a different prediction accuracy score, and because we were not interested in the robustness with respect to false predictions, we normalized the test data for correct predictions for each model to have a common ground to compare them amongst each other.
We did that by creating model specific test data from the CIFAR10 test set by selecting only those images that were correctly classified without any perturbation ($\epsilon=0$).
Thus, all models exhibit $100 \%$ accuracy in predictions on their individual test in the beginning.
On that basis, we measure the failure rate between models by assessing their accuracy decrease relative to the correctly predicted subset (relative accuracy) for different perturbation strengths $\epsilon$.
Amongst all adversarial methods, we have varied $\epsilon$ uniformly on a logarithmic scale until relative accuracy saturated.
Plotting relative accuracy against $\epsilon$ yields smooth curves, whereby a faster or steeper drop in relative accuracy corresponds to a less resilient classifier.
Next to this set-wide measure, we recorded the minimal fooling, or critical, distance $\|\Delta x\|_{\text{crit}}$ for each image in the individual test set and each classifier.
This measure corresponds to the Euclidean length of the perturbation vector in the input image space when the corresponding image is just misclassified.
The resulting distributions of critical distances and their statistics provide additional information with respect to the nature of resilience on the individual image level.

To get a qualitative impression on the representational geometry, we examine the model's decision landscape on a randomly projected plane in input space as in \cite{Hoffman.2019}.
In particular, we sample two random orthogonal directions in the input space that define a two-dimensional plane around a correctly classified image.
This allows one to plot the model's color-coded decisions $\argmax \left[ y(x) \right]$ in a neighborhood around a correct prediction.
To get an impression of the model's confidence, the contour lines of the maximum value of the softmax of $y(x)$ are superimposed.
In addition, we also visualize the map of the hidden representation's Jacobian norm $\| D\hat{h}(x) \|_{2}$ on the same plane which provides quantitative information about the local change of the encoding and an impression about its relative change, reflecting on the curvature of the surface.

\section{Results}\label{res}


\subsection*{Comparison in adversarial robustness}
\begin{figure}[ht]
\begin{center}
\includegraphics[width=\linewidth]{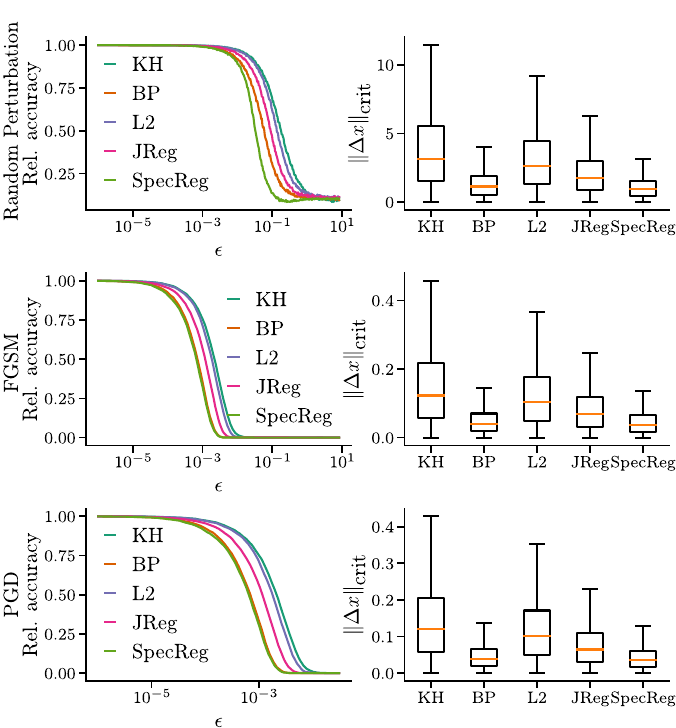}
\end{center}
\caption{\textbf{Left panel:} Relative accuracy as a function of the perturbation parameter $\epsilon$ for all models of consideration under the three adversarial attacks. \textbf{Right panel:} Distributions of critical perturbation magnitude $\|\Delta x\|_{\text{crit.}}$ as L2 distance in the input space (minimal fooling distance) across all images in the input set that were originally correctly classified for all models and attacks considered.}
\label{fig:AdversarialResults}
\end{figure}

We tested the robustness of Krotov and Hopfield's hybrid model (KH) against random perturbations in the input as well as FGSM and PGD adversarial attacks in comparison with end-to-end backpropagation trained models.
Next to a naive end-to-end SHLP model (BP), we tested L2, Jacobian (JReg)  \citep{Hoffman.2019} and spectral regularization (SpecReg) \citep{Nassar.2020}.
The results of our perturbation experiments are shown in Figure~\ref{fig:AdversarialResults}.
The left hand side panel show relative accuracy as a function of the perturbation parameter $\epsilon$.
Additionally, the right hand side panel visualizes the distributions of critical (minimum fooling) distances $\|\Delta x\|_{\text{crit}}$.
Across all attacks and measures, the hybrid model outperforms the others regarding robustness, followed by L2 and Jacobian regularization in that order.
In terms of FGSM and PGD resilience, the naive and the spectrally regularized model perform similarly bad.
However, concerning random perturbations, spectral regularization evidently yields worse results.
Generally across models, the order of severity is: random perturbations, FGSM and PGD.
Thus, robustness declines according to how specifically targeted the attack is.
As expected, relative accuracy, mean and median critical distance contain the same information.
Interestingly, however, the mean and the variance in critical distance appear to be directly correlated.
This means that with more resilient models, selected images are also less coherently correctly classified.
Consequently, the least robust models are also the most coherent.

\subsection*{Robustness-accuracy trade-off}
\begin{figure}[ht]
\begin{center}
\includegraphics[width=\linewidth]{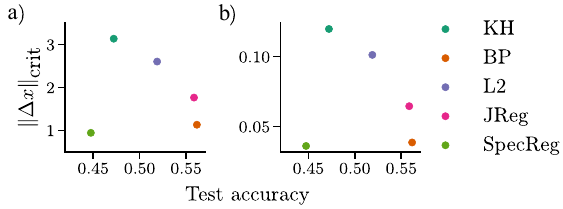}
\end{center}
\caption{
Relational plot between test accuracy and the median of critical distances $\|\Delta x\|_{\text{crit.}}$ across models as a measure of robustness regarding \textbf{a)}: random perturbations and \textbf{b)}: Projected Gradient Descent.
}
\label{fig:AccVSDcrit}
\end{figure}

We have studied the relation between robustness and accuracy in terms of test accuracy and median critical fooling distance $\|\Delta \mathbf{x} \|_{\text{crit.}}$ for all five models.
Figure~\ref{fig:AccVSDcrit} shows the relation between test accuracy and median critical fooling distance $\|\Delta x \|_{\text{crit.}}$ of all five models. 
As an example, we present results for a) random perturbations and b) PGD.
Aside from spectral regularization, robustness and accuracy are negatively correlated.
Although being the most robust, the hybrid model (KH) is also the least accurate.
The opposite applies to the naive model (BP).
Weight regularization in general finds a better balance compared to the other methods.
In particular, local weight regularization (JReg) achieves more robust representations without substantial decrease in accuracy, whereas L2 clearly favors robustness. 
Here, spectral regularization exhibits the worst results in both measures compared with the other models.
\label{robustness}

So far, we have observed a general trend between geometric regularization (L2, JReg) and robustness, compared to the hybrid and the naive model.
To categorize these results, we will study the mutual implications of geometric and spectral properties next.

\subsection*{Local learning yields optimal power law representations}
\begin{figure}[ht]
\begin{center}
\includegraphics[width=\linewidth]{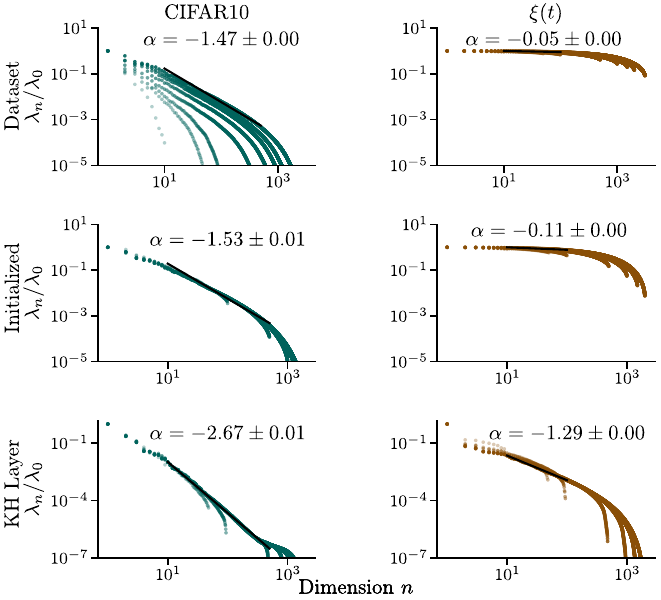}
\end{center}
\caption{Normalized covariance (PCA) spectra of latent representations $h(x)$ across CIFAR10 and Gaussian white noise $\xi(t)$ random input. The displayed spectra are those of the signals themselves, the hybrid model suggested by \cite{Krotov.2019} after initialization (Initialized) and after unsupervised training (KH Layer).}
\label{fig:LLSpectra}
\end{figure}

Figure~\ref{fig:LLSpectra} shows the ordered and normalized covariance spectra of the data themselves, the Krotov-Hopfield layer right after initialization (Initialized) and after training on CIFAR10 (KH Layer) on double logarithmic axes.
The spectra are simultaneously shown at different scales to examine their scaling behavior.
Our control, the white noise signal ($\xi(t)$), reveals a flat spectrum as expected besides the final fall due to the finite extensions of the model and data.
Also the latent representations of white noise just after initialization are flat.
At larger scales, the drop shifts towards the right, but the general flat profile of the spectrum is not affected by the scaling.
In turn, we can use the white noise spectra to calibrate our analysis of other spectra. 
For example, because the spectrum in the region appears completely flat, it is reasonable to have high confidence in spectrum profiles anywhere below $n \simeq 500$.
In addition, we note that linear regression yields exponents different from $0$, contrary to the real value. 
Therefore, we take the magnitude of this deviation as a proxy for the error of the estimate for $\alpha$.

CIFAR10 test images themselves do not exhibit a power law. 
Neither does the spectrum's profile appear linear, nor the spectral relation scale independent.
However, selecting sub-patches to scale the input dimension might have affected the integrity of the spectrum.
Surprisingly, the untrained network appears to have a scale free spectrum, even for CIFAR10 input.

The essential result of our spectral analysis is that the latent representations of the trained encoder consistently exhibit a scale free power law spectrum in the region $n < 800$.
Scaling only leads to earlier or later drops in the profile, but does not affect its overall shape.
Moreover, we notice a slight bump in CIFAR10 spectra for $n > 800$ dimensions which we account to finite boundary effects.
Surprisingly, even random signals appear to get projected to power law representations.
In general, we notice that the estimated exponents of representations are always larger than those of the pure data.
Moreover, exponents related to the CIFAR10 signal are larger than those related to the white noise source.

\subsection*{The link between geometry, spectrum and performance}
\begin{figure}[ht]
\begin{center}
\includegraphics[width=\linewidth]{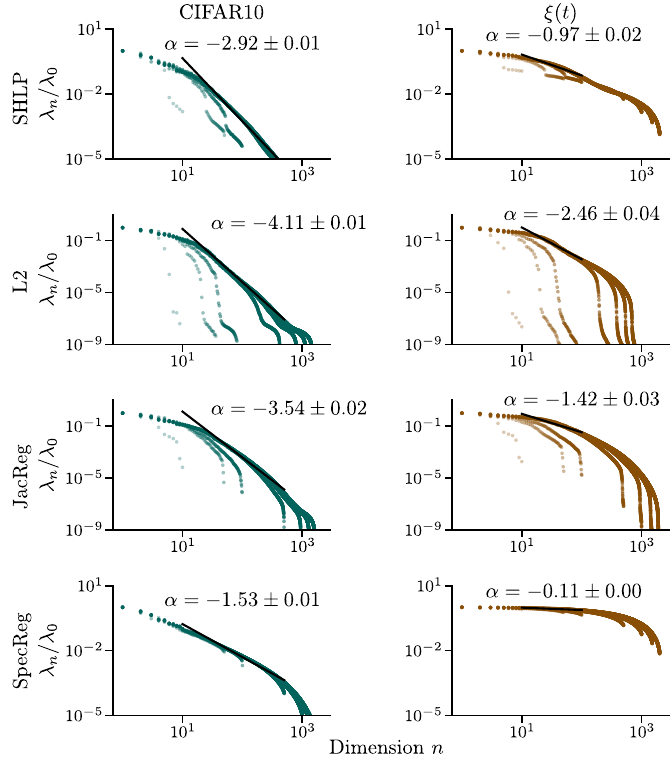}
\end{center}
\caption{Normalized covariance (PCA) spectra of latent representations $h(x)$ across CIFAR10 and Gaussian white noise $\xi(t)$ random input. The displayed spectra are those of the fully gradient optimised models without (SHLP) and with regularisation (L2, JacReg, SpecReg).}
\label{fig:ModelSpectra}
\end{figure}

Figure~\ref{fig:ModelSpectra} summarizes the ordered covariance spectra of all end-to-end backpropagation trained models for CIFAR10 as well as Gaussian white noise input.
At first, we note that the naive SHLP does not have a power spectrum, neither in terms of its profile shape nor its scaling.
The white noise spectrum seems to have a particularly complex profile.
L2 and Jacobian regularization appear to produce qualitatively similar spectra with steeper decay slopes than the naive model.
With respect to that, the L2 spectrum exhibits an even larger exponent estimate than that of the Jacobian regularizer.
It is interesting that weight regularization appears to cause a similar bump in the spectrum towards higher $n$ as in the KH Layer plot in Figure~\ref{fig:LLSpectra}.
The fact that this bump is more pronounced in the steeper L2 spectrum suggests that this might be a result of higher compression in the more dominant components.
As expected, spectral regularization achieves a good power law spectrum for CIFAR10 input.
In terms of quality, it is comparable to that of the hybrid model in Figure~\ref{fig:LLSpectra}.
However, since the model was optimized for $\alpha=1$, its estimated exponent including the inaccuracy deviates more from the target than expected.
Moreover, we see that for white noise input, the spectrally regularized model exhibits a completely flat spectrum, similar to the flat white noise spectra.
On closer inspection, we notice that both spectra resemble those of the untrained encoder in Figure~\ref{fig:LLSpectra}.
Overall, we observe that optimizing for smoothness in addition to accuracy does not seem sufficient to enforce power law spectra.
Additionally, the power law spectrum in SpecReg does not generalize to arbitrary inputs in contrast to the KH Layer.

\begin{figure}[ht]
\begin{center}
\includegraphics[width=\linewidth]{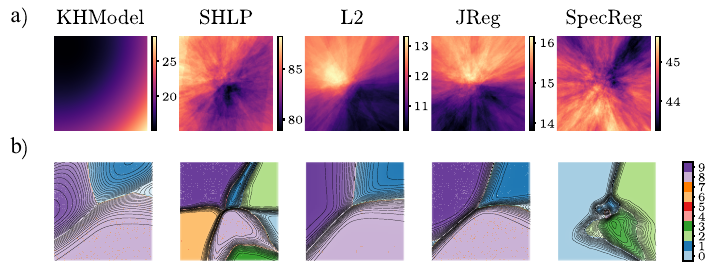}
\end{center}
\caption{Norm of the Jacobian of latent activations $\hat{\mathbf{h}}$ (upper panel) and decision landscape (lower panel) of studied models. Similarly to \cite{Hoffman.2019}, the displayed planes are created by grid sampling from a random plane projection in the model's input space.}
\label{fig:JacDecLandscapes}
\end{figure}

As in \cite{Hoffman.2019}, we plotted an exemplary decision boundary landscape of a common random plane projection in the model's input space. 
The resulting decision landscapes of all models are shown in the lower panel of Figure~\ref{fig:JacDecLandscapes}.
There, we started with one example image that was correctly classified across all models (center of the plots), and tracked the model's decisions along with its confidence to estimate the decision landscape for a linear continuation around the original image in a random two-dimensional plane.
In addition, we also visualize the Frobenius norm of the exact Jacobian of latent activations $\hat{h}$ of each model for the same random projection, as a measure of its smoothness.
With it, we gain two pieces of information within the plane.
The value of the norm serves as a local estimate of the encoding manifold's change.
Moreover, the relative change of the value of the norm in space provides a qualitative estimate of its curvature.
Thus, in total we get qualitative information concerning the latent and hidden representation's roughness, and how this affects the model's predictions, out of this plot.
In direct comparison, we observe that weight regularization (L2, JReg), in general, results in the lowest norm values, whereby L2 regularization achieves the lowest. 
Also, these low values in panel a), indicating comparably small changes in the representation surface, translate into smooth decision boundaries with very clear borders in panel b).
In contrast to that, the naive model (SHLP) exhibits by far the highest Jacobian scores which translate into rather small decision domains with rougher edges. 
Regarding the norm values, spectral regularization resides between those poles and shows rather large decision domain patches.
However, especially in the close neighborhood of the center, which represents the starting image, the decision terrain appears particularly steep.
In terms of the range of the Jacobian, Krotov and Hopfield's model (KHModel) is the closest to weight regularization, which is also reflected in its decision landscape.
In fact, the general structure of the decision lanscape between these models is almost identical, besides the orientation of the gradient.
However, the hybrid model's confidence changes less abruptly between decision domains, compared to the weight regularized models.

In terms of relative change of the Jacobian norm across the field, there are essential differences between local learning and backpropagation.
We observe that the norm value changes only gradually and smoothly along the latent activation landscape of the KH model. 
Moreover, the Jacobian score changes linearly from smaller to higher values starting in the upper left corner and ending in the lower right one.
These quantitative results indicate the highest curvature of the latent representation to occur perpendicular to the gradient which coincides with the major trench in the corresponding decision landscape.
From this, we see that the major trend in the geometry of the model's latent representation essentially informs the model's decision.
The same visual analysis yields similar results in case of L2 and Jacobian regularization.
However, we observe that the relative changes in the hidden Jacobian landscapes, which is reflected in the image's contrast, are less smooth which suggests a rougher surface.
Eventually, the respective relative changes in the naive and spectrally regularized models appear similarly abrupt and show no trend, particularly in the case of spectral regularization.

\section{Discussion}\label{disc}

In this paper, we studied the mutual implications between geometric and spectral properties of latent and hidden representations and how they affect performance of simple two layer perceptron models on the basis of Krotov and Hopfield's local learning model.
This theme resonates with the exploration undertaken in \cite{Wang.2021}, where the structural intricacies of Generative Adversarial Networks (GAN)' s latent spaces and their spectral properties are analyzed, elucidating their influence on image generation.
We measured performance in terms of accuracy and robustness against random perturbations, FGSM as well as PGD attacks.
In general, robustness and accuracy are consistently negatively correlated across models apart from spectral regularization, where the hybrid model was most robust but least accurate.
To understand why, we studied the smoothness of the model's representation manifold in terms of the Jacobian as well as its covariance spectrum with regard to power law profiles.
Both are established mechanisms to achieve general model robustness against data corruption.
To establish a baseline for comparison, we also studied the regularizers that optimize for the respective presumably optimal properties.
The local learning model exhibits both, a comparably smooth representation surface as well as a power law spectrum, indicating presumably optimally balanced representations.

Krotov and Hopfield's model yields latent representation manifolds similarly smooth to weight regularization.
Providing additional results on white box attacks, we find that smoother representation manifolds result in more resilient models in agreement with \cite{Krotov.2016, Krotov.2019, Grinberg.2019} and \cite{Hoffman.2019}.
By comparison with their decision landscapes, we see that geometric properties in hidden representations translate down to geometric properties in the classification layer, although they were not explicitely regularized.
With this, smoother hidden representations yield smoother decision boundaries, thus increasing robustness of the classifier overall.
This finding is in line with \cite{Zavatone-Veth.2023}, who explore how training induces geometric transformations in neural networks, particularly magnifying areas near decision boundaries, which significantly impacts class differentiation and network robustness.
Besides the local learning model, L2 as well as Jacobian regularization constitute the most promising approaches of those studied to achieve high resiliency.
Although they were optimized for accuracy and smoothness simultaneously, neither of the weight regularized models exhibit spectra close to a power law.
Consequently, we conclude that either both models are located afar from the optimum in parameter space. 
Moreover, an ideal balance might not be a sufficient criterion for this class of models.
Despite their spectral differences, we also observed similarities that were only found in the spectra of weight regularized models and the hybrid model.
In particular, we saw that all three featured a bump in their profiles for large $n$.
Since the synapse trajectories that are solutions to Krotov and Hopfield's learning rule approach weight bounded steady states, the hidden representations of the hybrid model are implicitly weight regularized.
Additionally, because L2 regularization also implies Jacobian regularization (Section~\ref{meth}), this similarity explains the common characteristics between the three models.
Overall, we conclude that the hybrid model's robustness can, in part, be explained by smooth representations and weight regularization.
The latter also correlates with the high frequency bump characteristic in the spectra.

Controlling the spectrum directly had almost no implications, neither regarding geometry nor performance.
Following \cite{Nassar.2020}, we would have expected an increase in robustness from spectral regularization that we did not reproduce.
We observe that spectral regularization decreases the magnitude of the latent Jacobian compared to the naive representation but does not benefit robustness 
This can be explained by a more strongly folded surface, which the abrupt changes in the Jacobian norm hint at.
In contrast to local learning, the regularized power spectrum does not generalize to white noise data.
Consequently, the hybrid model constitutes the more interesting case to study.
Its latent spectrum falls more quickly than the dataset's which can be seen from the estimated exponents.
If we assume them to reflect the intrinsic (fractal) dimension of the signal, even when the spectra do not follow power laws, we qualitatively confirm \cite{Stringer.2019} in that optimal representations have higher exponents corresponding to lower dimensions.
However, representations are also generally expected to are lower dimensional than the original data because they formally constitute some form of data compression.
This argument also explains the series of decay exponents that where estimated in Figure~\ref{fig:LLSpectra}.
The randomly initialized model exhibits the flattest spectrum followed by the dataset.
Eventually, the spectrum of learned representations decays significantly faster which indicates that the model has learned to structure the data.
In this light, also the estimated exponents of the end-to-end backpropagated models are consistent.
For example, the decay of the weight regularized models is steeper in the estimation regions compared to the naive model since constraint (smoother) representations lead to higher degrees of compression.
In particular, the L2 spectrum is characterized by an even steeper fall in comparison to Jacobian regularization.
As predicted by \cite{Stringer.2019}, the spectral decay is generally stimulus dependent, with flatter spectra reflecting higher dimensional data.

Although robustness of the hybrid model can in part be understood by weight regularization, we notice that there remain gaps that are not explained by the current state of the theory.
Partly, they might be a result of our model's finite nature whereas Stringer's arguments rely on properties in infinitely dimensional Hilbert spaces.
In any case, our results suggest that many open questions remain regarding understanding robustness of classifiers, even for simple function classes.

To close these gaps, our discovery, that Krotov and Hopfield's local learning rule yields robust representations that perform well, are smooth and exhibit a close to ideal power law spectrum, might be of significant impact for upcoming studies.
With properties that match with the ideal model in Stringer's theory it is a promising mechanistic study case.
Moreover, our work could provide a starting point towards understanding how power law spectra determine optimally smooth encodings and beyond.
Because of the model's biological foundations, our results also provide insights into how robust neural networks are mechanistically realized in the mammalian brain, and how they can be achieved in artificial systems.

\backmatter

\section*{Declarations}

\subsection*{Funding}
This research was funded by the Research Council of Norway Grant 300504, the University of Oslo and Simula Research Laboratory.

\subsection*{Conflict of interest}
The authors declare no competing interests.

\subsection*{Ethics approval and consent to participate}
Not applicable

\subsection*{Consent for publication}
Not applicable

\subsection*{Materials availability}
Not applicable

\subsection*{Data and Code Availability}

In alignment with the submission guidelines and commitment to ensuring a rigorous double-blind review process, the data supporting the findings of this study, along with the corresponding code and detailed instructions for reproducing the results, are available. 
The repository includes all necessary model files, data and code used for generating the figures presented within the manuscript, and the figures themselves. 
This collection can be accessed through a public and referable Zenodo archive and the GitHub repository linked therein \citep{zenodo_repository_archive}.
The repository is structured to facilitate easy navigation and replication of the study's findings. It includes a README file that provides instructions for replicating the analyses and figures as an independent sample. 
This approach ensures that our data is fully transparent and accessible, adhering to the principles of open science, while maintaining the integrity of the double-blind review process.
Our results were generated using the methods we mention with parameters according to their references.
Whenever our parameter values differ from that in the resources we explicitly state our values in the text.
Upon the publication of this study, the repository is made publicly available under the authors' names, ensuring full transparency and accessibility of the research materials in line with open science principles.

\subsection*{Author contribution}

\begin{itemize}[label={}, leftmargin=*]
  \item \textbf{Konstantin Holzhausen}: Conceived and designed the study, wrote code, performed simulations, analyzed and interpreted results, and wrote the manuscript.
  
  \item \textbf{Mia Merlid}: Contributed to the study design, wrote substantial parts of the code, performed simulations, collected results, and assisted in drafting the manuscript.
  
  \item \textbf{Håkon Olav Torvik}: Contributed to the study design, performed simulations, collected, analyzed, and interpreted results, and provided essential intellectual input.
  
  \item \textbf{Anders Malthe-Sørenssen}: Contributed to study conception, provided supervision, drafted the manuscript, and offered intellectual input.
  
  \item \textbf{Mikkel Elle Lepperød}: Conceived the original idea, provided supervision and guidance, wrote code, contributed to drafting and writing the manuscript, and provided essential intellectual input and critical feedback.
\end{itemize}

\bibliography{bibliography}

\end{document}